\documentclass[11pt,a4paper]{article}
\PassOptionsToPackage{breaklinks}{hyperref}
\usepackage[hyperref]{acl2019}
\usepackage[utf8]{inputenc}
\usepackage{times}
\usepackage{latexsym}
\usepackage{pifont}

\usepackage{graphicx}
\usepackage{multirow}
\usepackage{tabularx}
\usepackage{breakurl}
\usepackage{caption}
\usepackage{subcaption}
\usepackage{xspace}
\usepackage{pifont}

\usepackage{textcomp}

\newenvironment{citemize}{\begin{list}{$\bullet$}{\topsep=\smallskipamount\itemsep=0pt\parsep=1pt\labelwidth=.5em}}{\end{list}}

\newcommand{\udpipe}{\emph{UDPipe 2.0}\xspace}
\newcommand{\udst}{\emph{CoNLL 2018 Shared Task: Multilingual Parsing from Raw Text to Universal Dependencies}\xspace}
\newcommand{\epe}{\emph{The 2018 Shared Task on Extrinsic Parser Evaluation}\xspace}
\newcommand{\sigm}{\emph{SIGMORPHON 2019}\xspace}
\newcommand{\sigmorphon}{\emph{SIGMORPHON 2019 Shared Task: Crosslinguality and Context in Morphology}\xspace}

\newcommand{\YES}{\ding{51}\xspace}
\newcommand{\NO}{{\color{gray!50}\ding{55}}\xspace}

\aclfinalcopy
\def\aclpaperid{***} 

\title{UDPipe at SIGMORPHON 2019: Contextualized Embeddings,
Regularization with Morphological Categories, Corpora Merging}

\author{Milan Straka \and Jana Strakov\'{a} \and Jan Haji\v{c}\\
  Charles University \\
  Faculty of Mathematics and Physics \\
  Institute of Formal and Applied Linguistics \\
  {\tt \{straka,strakova,hajic\}@ufal.mff.cuni.cz} \\}

\date{}

\begin{document}
\maketitle

\begin{abstract}
  We present our contribution to the \sigmorphon, Task 2: contextual
  morphological analysis and lemmatization.

  We submitted a modification of the \udpipe, one of best-performing systems of
  the \udst and an overall winner of the \epe.

  As our first improvement, we use the pretrained contextualized embeddings
  (BERT) as additional inputs to the network; secondly, we use individual
  morphological features as regularization; and finally, we merge the selected
  corpora of the same language.

  In the lemmatization task, our system exceeds all the submitted systems by
  a wide margin with lemmatization accuracy $95.78$ (second best was $95.00$,
  third $94.46$). In the morphological analysis, our system placed tightly
  second: our morphological analysis accuracy was $93.19$, the winning system's
  $93.23$.
\end{abstract}

\section{Introduction}
\label{section:introduction}

This work describes our participant system in the \sigmorphon. We contributed
a system in Task 2: contextual morphological analy\-sis and lemmatization.

Given a segmented and tokenized text in a CoNLL-U format with surface forms
(column 2) as in the following example:

\begin{small}
\begin{verbatim}
# sent-id = 1
# text = They buy and sell books.
1  They   _     _  _  _           ...
2  buy    _     _  _  _           ...
3  and    _     _  _  _           ...
4  sell   _     _  _  _           ...
5  books  _     _  _  _           ...
6  .      _     _  _  _           ...
\end{verbatim}
\end{small}

\noindent the task is to infer lemmas (column 3) and morphological analysis
(column 6) in the form of concatenated morphological features:

\begin{small}
\begin{verbatim}
# sent-id = 1
# text = They buy and sell books.
1  They   they  _  _  N;NOM;PL    ...
2  buy    buy   _  _  V;SG;1;PRS  ...
3  and    and   _  _  CONJ        ...
4  sell   sell  _  _  V;PL;3;PRS  ...
5  books  book  _  _  N;PL        ...
6  .      .     _  _  PUNCT       ...
\end{verbatim}
\end{small}

The \sigm data consists of $66$ distinct languages in $107$ corpora
\citep{mccarthy-etal-2018-marrying}.

We submitted a modified \udpipe~\cite{UDPipe2.0}, one of the three winning
systems of the \udst~\cite{CoNLL2018} and an overall winner of the
\epe~\cite{EPE2018}.

Our improvements to the \udpipe~are threefold:

\begin{citemize}
  \item We use the pretrained contextualized embeddings (BERT) as additional
    inputs to the network (described in Section~\ref{section:bert}).
  \item Apart from predicting the whole POS tag, we regularize the model
    by also predicting individual morphological features
    (Section~\ref{section:feature_regularization}).
  \item In some languages, we merge all the corpora of the same language (Section~\ref{section:corpora_merging}).
\end{citemize}

Our system placed first in lemmatization and closely second in morphological
analysis.

We give an overview of the related work in Section~\ref{section:related_work},
we describe our methodology in Section~\ref{section:methods}, the results with
ablation experiments are given in Section~\ref{section:results} and we conclude in
Section~\ref{section:conclusions}.

\section{Related Work}
\label{section:related_work}

A new type of deep contextualized word representation was introduced by
\citet{Peters2018}. The proposed embeddings, called ELMo, were obtained from
internal states of deep bidirectional language model, pretrained on a large
text corpus. The idea of ELMos was extended by \citet{BERT}, who instead of
a bidirectional recurrent language model employ a Transformer
\citep{vaswani:2017} architecture.

The \emph{Universal
Dependencies} 
project \citep{ud} seeks to develop cross-linguistically consistent treebank
annotation of morphology and syntax for many languages.
In 2017 and 2018 CoNLL Shared Tasks \emph{Multilingual Parsing from Raw Text
to Universal Dependencies} \cite{CoNLL2017,CoNLL2018}, the goal was to process raw
texts into tokenized sentences with POS tags, lemmas, morphological features
and dependency trees of Universal Dependencies. \citet{UDPipe2.0} was one of the
winning systems of the 2018 shared task, performing the POS tagging,
lemmatization and dependency parsing jointly. Another winning system of
\citet{udst18:harbin} employed manually trained ELMo-like contextual word
embeddings and ensembling, reporting $7.9\%$ error reduction in LAS parsing performance.

The Universal Morphology (UniMorph) is also a project seeking to provide
annotation schema for morphosyntactic details of language \citep{UniMorph}. Each POS tag
consists of a set of morphological features, each belonging to a morphological
category (also called a dimension of meaning).

\section{Methods}
\label{section:methods}

\begin{figure}[t]
  \begin{center}
    \includegraphics[width=\hsize]{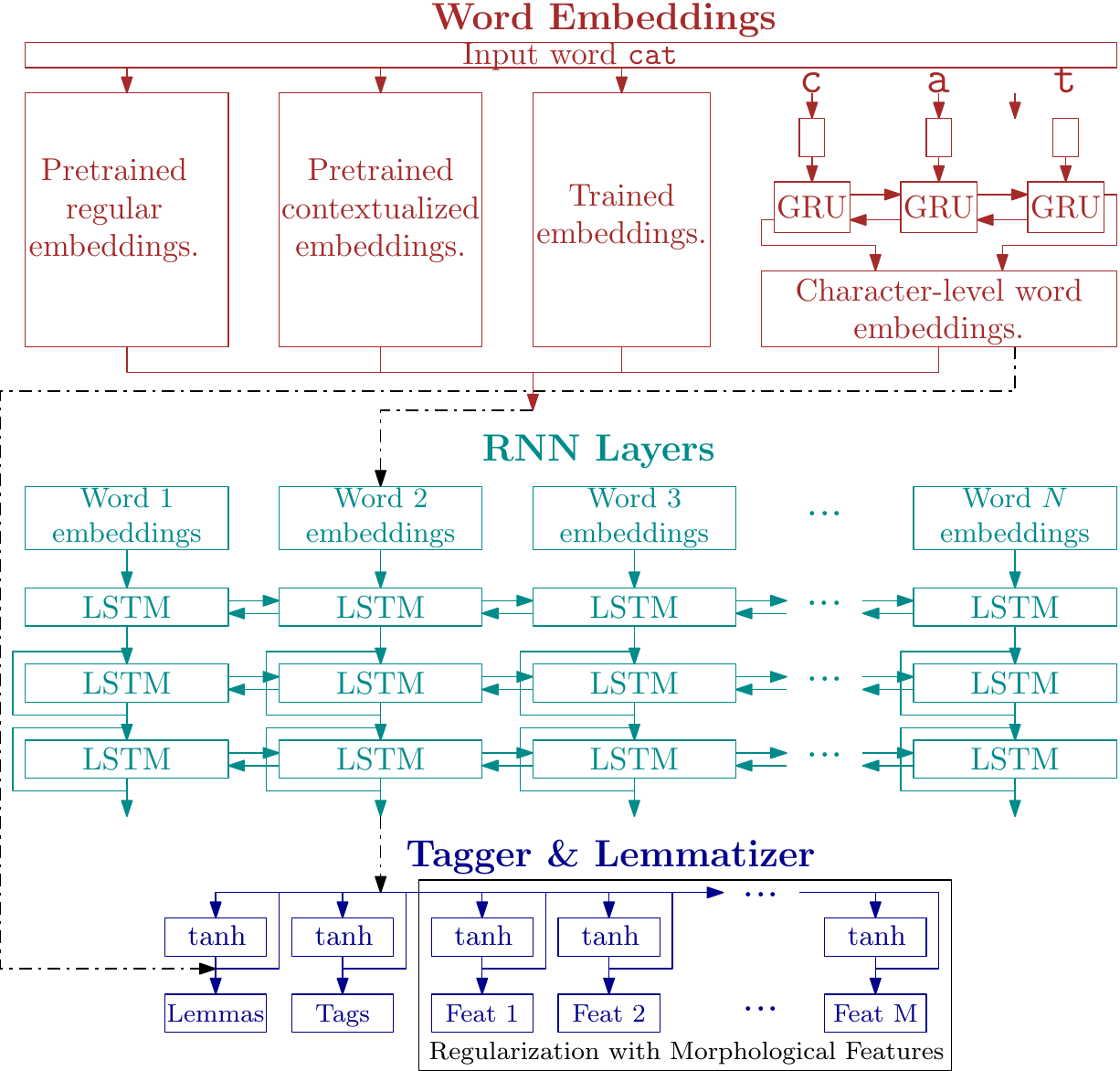}
  \end{center}
  \caption{The overall system architecture}
  \label{fig:architecture}
\end{figure}

\subsection{Architecture Overview}

\begin{table*}[t]
  \begin{center}
    \small
    \begin{tabular}{l||l|l||l}
      \multicolumn{1}{c||}{Lemma Rule} & \multicolumn{1}{c|}{Casing Script}
        & \multicolumn{1}{c||}{Edit Script} & \multicolumn{1}{c}{Most Frequent Examples} \\\hline\hline
        \verb|↓0;d¦|      & all lowercase           & do nothing            & the→the to→to and→and \\\hline
        \verb|↑0¦↓1;d¦|   & first upper, then lower & do nothing            & Bush→Bush Iraq→Iraq Enron→Enron \\\hline
        \verb|↓0;d¦-|     & all lowercase           & remove last character & your→you an→a years→year\\\hline
        \verb|↓0;abe|     & all lowercase           & ignore form, use \verb|be| & is→be was→be 's→be\\\hline
        \verb|↑0;d¦|      & all uppercase           & do nothing            & I→I US→US NASA→NASA\\\hline
        \verb|↓0;d¦--|    & all lowercase           & remove last 2 chars   & been→be does→do called→call\\\hline
        \verb|↓0;d¦---|   & all lowercase           & remove last 3 chars   & going→go being→be looking→look\\\hline
        \verb|↓0;d--+b¦|  & all lowercase           & change first 2 chars to \verb|b| & are→be 're→be Are→be\\\hline
        \verb|↓0;d¦-+v+e| & all lowercase           & change last char to \verb|ve| & has→have had→have Has→have\\\hline
        \verb|↓0;d¦---+e| & all lowercase           & change last 3 chars to \verb|e| & having→have using→use making→make\\\hline
        \verb|↓0;d¦-+o→|  & all lowercase           & change last but 1 char to \verb|o| & n't→not knew→know grew→grow\\
    \end{tabular}
  \end{center}
  \caption{Eleven most frequent lemma rules in English EWT corpus, ordered from the most frequent one.}
  \label{table:lemmarules}
\end{table*}

Our \textbf{baseline} is the \udpipe~\cite{UDPipe2.0}. The original \udpipe is
available at \url{http://github.com/CoNLL-UD-2018/UDPipe-Future}. Here, we
describe the overall architecture, focusing on the modifications made for the
\sigm. The resulting model is presented in Figure~\ref{fig:architecture}.

In short, \udpipe is a multi-task model predicting POS tags, lemmas and
dependency trees. For the \sigm, we naturally train and predict only the POS
tags (morphosyntactic features) and lemmas. After embedding input words, three
shared bidirectional LSTM~\citep{Hochreiter:1997:LSTM} layers are performed.
Then, softmax classifiers process the output and generate the lemmas and POS
tags (morphosyntactic features).

The lemmas are generated by classifying into a set of edit scripts which
process input word form and produce lemmas by performing character-level edits
on the word prefix and suffix. The lemma classifier additionally takes the
character-level word embeddings as input. The lemmatization is further
described in Section~\ref{section:lemmatization}.

The input word embeddings are the same as in the \udpipe~\cite{UDPipe2.0}:

\begin{citemize}
  \item \textbf{end-to-end word embeddings},
  \item \textbf{word embeddings (WE):} We use FastText word embeddings
    \cite{FastText} of dimension $300$, which we pretrain for each language
    on plain texts provided by CoNLL 2017 UD Shared Task, using segmentation and
    tokenization trained from the UD data.\footnote{We use
    {\scriptsize\ttfamily-minCount 5 -epoch 10 -neg 10} options.} For languages
    not present in the CoNLL 2017 UD Shared Task, we use pretrained embeddings
    from \cite{crawlvectors}, if available.
  \item \textbf{character-level word embeddings (CLE):} We employ bidirectional
    GRUs \cite{Cho2014,Graves2005} of dimension $256$ in line with
    \cite{Ling2015}: we represent every Unicode character with a vector of
    dimension $256$, and concatenate GRU output for forward and reversed word
    characters. The character-level word embeddings are trained together with
    UDPipe network.
\end{citemize}

We refer the readers for detailed description of the architecture and the
training procedure to \citet{UDPipe2.0}.

The main modifications to the \udpipe are the following:

\begin{citemize}
  \item \textbf{contextualized embeddings (BERT)}: We add pretrained contextual
    word embeddings as another input to the neural network. We describe this
    modification in Section~\ref{section:bert}.
  \item \textbf{regularization with individual morphological features}: We
    predict not only the full POS tag, but regularize the model by also predicting
    individual morphological features, which is described in
    Section~\ref{section:feature_regularization}.
  \item \textbf{corpora merging}: In some cases, we merge the corpora of the
    same language. We describe this step in
    Section~\ref{section:corpora_merging}.
\end{citemize}

Furthermore, we also employ model ensembling, which we describe in
Section~\ref{section:model_ensembling}.

\subsection{Lemmatization}
\label{section:lemmatization}

The lemmatization is modeled as a multi-class classification, in which the
classes are the complete rules which lead from input forms to the lemmas. We
call each class encoding a transition from input form to lemma a \textit{lemma
rule}. We create a lemma rule by firstly encoding the correct casing as
a \textit{casing script} and secondly by creating a sequence of character
edits, an \textit{edit script}.

Firstly, we deal with the casing by creating a \textit{casing script}. By
default, word form and lemma characters are treated as lowercased. If the lemma
however contains upper-cased characters, a rule is added to the casing script
to uppercase the corresponding characters in the resulting lemma. For example,
the most frequent casing script is ``keep the lemma lowercased (don't do
anything)'' and the second most frequent casing script is ``uppercase the first
character and keep the rest lowercased''.

As a second step, an \textit{edit script} is created to convert input
lowercased form to lowercased lemma. To ensure meaningful editing, the form is split to three parts,
which are then processed separately: a prefix, a root (stem) and a suffix. The
root is discovered by matching the longest substring shared between the form
and the lemma; if no matching substring is found (e.g., form \textit{went} and
lemma \textit{go}), we consider the word irregular, do not process it with any
edits and directly replace the word form with the lemma. Otherwise, we proceed
with the edit scripts, which process the prefix and the suffix separately and
keep the root unchanged. The allowed character-wise operations are character
copy, addition and deletion.

The resulting \textit{lemma rule} is a concatenation of a casing script and an
edit script. The most common lemma rules in English EWT corpus are presented
in Table~\ref{table:lemmarules}, and the number of lemma rules for every
language is displayed in
Tables~\ref{table:results-per-corpus-I}~and~\ref{table:results-per-corpus-II}.

Using the generated lemma rules, the task of lemmatization is then reduced to
a multiclass classification task, in which the artificial neural network
predicts the correct lemma rule.

\subsection{Contextual Word Embeddings (BERT)}
\label{section:bert}

We add pretrained contextual word embeddings as another input to the neural
network. We use the pretrained contextual word embeddings called BERT
\cite{BERT}.\footnote{\url{https://github.com/google-research/bert}} For
English, we use the native English model ({\small\ttfamily BERT-Base English}),
for Chinese use use the native Chinese model ({\small\ttfamily BERT-Base
Chinese}) and for all other languages, we use the Multilingual model
({\small\ttfamily BERT-Base Uncased}). All models provide contextualized
embeddings of dimension $768$.

We average the last four layers of the BERT model to produce the embeddings.
Because BERT utilizes word pieces, we decompose words into appropriate
subwords and then average the generated embeddings over subwords belonging to
the same word.

Contrary to finetuning approach used by the BERT authors \citep{BERT}, we never
finetune the embeddings.

\subsection{Regularization with Individual Morphological Features}
\label{section:feature_regularization}

Our model predicts the POS tags as a unit, i.e., the whole set of morphological
features at once. There are other possible alternatives -- for example, we
could predict the morphological features individually. However, such
a prediction needs to decide which morphological categories to use and should
use a classifier capable of handling dependencies between the predicted
features, and all our attempts to design such a classifier resulted in systems
with suboptimal performance. Using a whole-set classifier alleviates the need
for finding a correct set of categories for a word and handling the feature
dependencies, but suffers from the curse of dimensionality, especially on
smaller corpora with richer morphology.

Nevertheless, the performance of a whole-set classifier can be improved by
regularizing with the individual morphological feature prediction.
Similarly to \citet{LemmaTag}, our model predicts not only the full set
of morphological features at once, but also the individual features.
Specifically, we employ as many additional softmax output layers as the number of
morphological categories used in the corpus, each predicting the
corresponding feature or a special value of \texttt{None}. The averaged
cross-entropy loss of all predicted categories multiplied by a weight $w$ is
added to the training loss. The predicted features are not used in any way
during inference and act only as model regularization.

The number of full POS tags (complete sets of morphological features),
individual morphological features and number of used morphological categories
for every corpus is provided in
Tables~\ref{table:results-per-corpus-I}~and~\ref{table:results-per-corpus-II}.

\subsection{Corpora Merging}
\label{section:corpora_merging}

Given that the shared task data consists of multiple corpora for some of the
languages, it is a natural approach to concatenate all corpora of the same
language and use the resulting so-called \emph{merged model} for prediction on individual
corpora.

In theory, concatenating all corpora of the same language should be
always beneficial considering the universal scheme used for annotation.
Nonetheless, the merged model exhibits worse performance in many cases,
compared to a specialized model trained on the corpus training data only, supposedly
because of systematically different annotation. We consequently improve the merged
model performance during inference by allowing only such lemma rules and
morphological feature sets that are present in the training data of the
predicted corpus.

\subsection{Model Ensembling}
\label{section:model_ensembling}

For every corpus, we consider three model configurations -- the regular model with
BERT embeddings trained only on the corpus data, the merged model with BERT
embeddings trained on all corpora of the corresponding language, and the
no-BERT model trained only on the corpus data.

To allow automatic model selection and to obtain highest performance, we
use ensembling. Namely, we train three models for every model
configuration, obtaining nine models for every language. Then, we choose
a model subset whose ensemble achieves the highest performance on the
development data. The chosen subsets then formed the competition entry of our system.

However, post-competition examination using half of development data for
ensemble selection and the other for evaluation revealed that the model
selection can overfit, sometimes choosing one or two models with high
performance caused by noise instead of high-quality generalization. Therefore,
we also consider another model selection method -- we ensemble the three models
for every configuration, and choose the best configuration out of three
ensembles on the development data. This second system has been submitted as
a~post-competition entry.

\section{Results}
\label{section:results}

\begin{table*}
  \catcode`@ = 13\def@{\bfseries}
  \begin{center}
  \begin{tabular}{cc}

    \begin{tabular}{lc}
      \hline\multicolumn{2}{c}{Lemma Accuracy} \\ \hline
      @UFALPRAGUE-01      &@95.78 \\
      CHARLES-SAARLAND-02 & 95.00 \\
      ITU-01              & 94.46 \\
      baseline-test-00    & 94.17 \\
      CBNU-01             & 94.07 \\
    \end{tabular}

    &

    \begin{tabular}{lc}
      \hline\multicolumn{2}{c}{Morph Accuracy} \\\hline
      CHARLES-SAARLAND-02       & 93.23 \\
      @UFALPRAGUE-01            &@93.19 \\
      RUG-01                    & 90.53 \\
      EDINBURGH-01              & 88.93 \\
      RUG-02                    & 88.80 \\
    \end{tabular}

    \\

    \begin{tabular}{lc}
      \hline\multicolumn{2}{c}{Lemma Levenshtein} \\\hline
      @UFALPRAGUE-01            &@0.098 \\
      CHARLES-SAARLAND-02       & 0.108 \\
      ITU-01                    & 0.108 \\
      CBNU-01                   & 0.127 \\
      baseline-test-00          & 0.129 \\
    \end{tabular}

    &

    \begin{tabular}{lc}
      \hline\multicolumn{2}{c}{Morph F1} \\\hline
      CHARLES-SAARLAND-02       & 96.02 \\
      @UFALPRAGUE-01            &@95.92 \\
      RUG-01                    & 94.54 \\
      RUG-02                    & 93.22 \\
      EDINBURGH-01              & 92.89 \\
    \end{tabular}

    \\
  \end{tabular}
  \caption{Top 5 results in lemma accuracy, lemma Levenshtein,
  morphological accuracy and morphological F1.}
  \label{table:top5}
  \end{center}
\end{table*}

\begin{table*}
  \begin{center}
    \begin{tabular}{l|l|l||rr|rr}
      \multicolumn{1}{c|}{Word} & \multicolumn{1}{c|}{\multirow{2}{*}{BERT}} & \multicolumn{1}{c||}{Feature}
        & \multicolumn{2}{c|}{Lemma} & \multicolumn{2}{c}{Morph} \\\cline{4-7}
      \multicolumn{1}{c|}{Embeddings} & & \multicolumn{1}{c||}{Regularization}
        & \multicolumn{1}{c}{Acc} & \multicolumn{1}{c|}{Lev} & \multicolumn{1}{c}{Acc} & \multicolumn{1}{c}{F1} \\\hline\hline
      \NO     & \NO  & \NO          & 94.251 & 0.168 & 90.506 & 93.585 \\\hline
      \YES FT only & \NO  & \NO          & 95.229 & 0.109 & 91.704 & 94.745 \\
      \YES & \NO  & \NO          & 95.294 & 0.107 & 91.828 & 94.849 \\\hline
      \NO  & \YES & \NO          & 95.440 & 0.106 & 92.789 & 95.614 \\
      \YES & \YES & \NO          & 95.534 & 0.104 & 92.980 & 95.755 \\\hline
      \NO  & \NO  & \YES $w=1$   & 95.120 & 0.111 & 91.468 & 94.672 \\
      \YES & \NO  & \YES $w=1$   & 95.365 & 0.104 & 92.135 & 95.189 \\
      \YES & \YES & \YES $w=1$   & 95.516 & 0.105 & 93.148 & 95.957 \\
      \YES & \YES & \YES $w=0.5$ & 95.534 & 0.105 & 93.172 & 95.939 \\
      \YES & \YES & \YES $w=2$   & 95.539 & 0.105 & 93.175 & 95.965 \\\hline

    \end{tabular}
  \end{center}
  \caption{Lemma accuracy, lemma Levenshtein,
  morphological accuracy, and morphological F1 results of ablation experiments.
  For comparison, the \emph{FT only} embeddings denote the pretrained
  embeddings of \cite{crawlvectors}.}
  \label{table:ablations}
\end{table*}

\begin{table*}
  \begin{center}
    \begin{tabular}{l|l|l|l||rr|rr}
      \multicolumn{1}{c|}{Regular} & \multicolumn{1}{c|}{Merged} & \multicolumn{1}{c|}{Without} & \multicolumn{1}{c||}{\multirow{2}{*}{Ensembling}}
        & \multicolumn{2}{c|}{Lemma} & \multicolumn{2}{c}{Morph} \\\cline{5-8}
      \multicolumn{1}{c|}{Model} & \multicolumn{1}{c|}{Model} & \multicolumn{1}{c|}{BERT} &
        & \multicolumn{1}{c}{Acc} & \multicolumn{1}{c|}{Lev} & \multicolumn{1}{c}{Acc} & \multicolumn{1}{c}{F1} \\\hline\hline
      \YES & \NO     & \NO  & \NO        & 95.516 & 0.105 & 93.148 & 95.957 \\
      \YES & \YES    & \NO  & \NO        & 95.702 & 0.101 & 93.322 & 96.081 \\
      \YES & \NO     & \YES & \NO        & 95.524 & 0.104 & 93.177 & 95.966 \\
      \YES & \YES    & \YES & \NO        & 95.709 & 0.100 & 93.353 & 96.090 \\\hline
      \YES & \NO     & \NO  & \YES Every model       & 95.606 & 0.102 & 93.257 & 95.997 \\
      \YES & \YES    & \NO  & \YES configuration     & 95.785 & 0.098 & 93.422 & 96.123 \\
      \YES & \NO     & \YES & \YES has independent   & 95.598 & 0.102 & 93.300 & 96.035 \\
      \YES & \YES    & \YES & \YES 3-model ensemble  & 95.776 & 0.099 & 93.464 & 96.160 \\\hline\hline
      \multicolumn{8}{c}{The competition entry, which allows ensembling any combination of the 9 models}\\\hline
      \YES & \YES    & \YES & \YES Any combination   & 95.776 & 0.098 & 93.186 & 95.924 \\
    \end{tabular}
  \end{center}
  \caption{Lemma accuracy, lemma Levenshtein, morphological accuracy, and
  morphological F1 results of model combinations. When not specified otherwise,
  all models utilize pretrained word embeddings, BERT, and feature regularization with
  weight $w=1$.}
  \label{table:combination}
\end{table*}

\subsection{\sigm Test Results}

Table~\ref{table:top5} shows top 5 results in lemma accuracy, lemma
Levenshtein, morphological accuracy and morphological F1 in Task 2 of the
\sigm, averaged over all $107$ corpora. Our system is called UFALPRAGUE-01.

Our participant system placed as one of the winning systems of the shared task.
In the lemmatization task, our system exceeds all the submitted systems by
a wide margin with lemmatization accuracy $95.78$ (second best was $95.00$,
third $94.46$). In the morphological analysis, our system placed tightly second:
our morphological analysis accuracy was $93.19$, the winning system's $93.23$.

\begin{table*}
  \begin{center}
    \small
    \tabcolsep=1.9pt
    \begin{tabular}{l|r|rrr||rr|rr|ccc||rr|rr|r}
      \multicolumn{1}{c|}{\multirow{2}{*}{Treebank}} &
      \multicolumn{1}{c|}{\multirow{2}{*}{Words}} &
      \multicolumn{1}{c}{\kern-2ptLemma\kern-2pt} &
      \multicolumn{1}{c}{POS} &
      \multicolumn{1}{c||}{Feats/} &
      \multicolumn{2}{c|}{Lemma} & \multicolumn{2}{c|}{Morph} &
      \multicolumn{3}{c||}{Model} &
      \multicolumn{2}{c|}{Merged $\Delta$} &
      \multicolumn{2}{c|}{BERT $\Delta$} &
      \multicolumn{1}{c}{B} \\
      &&\multicolumn{1}{c}{Rules} &
      \multicolumn{1}{c}{Tags} &
      \multicolumn{1}{c||}{/Cats} &
      \multicolumn{1}{c}{Acc} & \multicolumn{1}{c|}{Lev} & \multicolumn{1}{c}{Acc} & \multicolumn{1}{c||}{F1} &
      \multicolumn{1}{c}{R} &
      \multicolumn{1}{c}{M} &
      \multicolumn{1}{c||}{N} &
      \multicolumn{1}{c}{LAcc} &
      \multicolumn{1}{c|}{\kern-2ptMAcc} &
      \multicolumn{1}{c}{LAcc} &
      \multicolumn{1}{c|}{\kern-2ptMAcc} &
      \multicolumn{1}{c}{T} \\\hline\hline
Afrikaans-AfriBooms & 38\,843 & 185 & 41 & 29/~9~ & 99.10 & 0.01 & 99.26 & 99.40 & \YES & \NO & \NO &  &  & 0.15 & 0.64 & \YES\\\hline
Akkadian-PISANDUB & 1\,425 & 548 & 12 & 11/~1~ & 55.94 & 1.50 & 86.63 & 86.46 & \NO & \NO & \YES &  &  & 0.99 & -4.45 & \NO\\\hline
Amharic-ATT & 7\,952 & 1 & 54 & 35/10 & 100.00 & 0.00 & 89.70 & 93.24 & \YES & \NO & \NO &  &  & 0.00 & 0.57 & \YES\\\hline
Ancient Greek-PROIEL & 171\,478 & 6\,843 & 887 & 49/14 & 94.04 & 0.15 & 92.99 & 97.92 & \NO & \YES & \NO & 0.65 & 0.30 & -0.04 & -0.01 & \NO\\\hline
Ancient Greek-Perseus & 162\,164 & 9\,088 & 795 & 44/11 & 91.91 & 0.21 & 91.91 & 96.74 & \NO & \YES & \NO & 0.78 & 0.26 & -0.05 & 0.01 & \NO\\\hline
Arabic-PADT & 225\,494 & 3\,174 & 287 & 35/12 & 96.09 & 0.11 & 95.38 & 97.48 & \YES & \NO & \NO & -0.09 & 0.02 & 0.35 & 0.66 & \YES\\\hline
Arabic-PUD & 16\,845 & 3\,919 & 371 & 39/12 & 79.26 & 0.78 & 86.52 & 95.30 & \YES & \NO & \NO &  &  & 0.46 & 3.22 & \YES\\\hline
Armenian-ArmTDP & 18\,595 & 277 & 360 & 65/16 & 95.91 & 0.08 & 93.63 & 96.54 & \YES & \NO & \NO &  &  & 0.62 & 1.90 & \YES\\\hline
Bambara-CRB & 11\,205 & 311 & 43 & 32/~9~ & 92.10 & 0.12 & 94.00 & 95.62 & \YES & \NO & \NO &  &  & -0.84 & 0.07 & \NO\\\hline
Basque-BDT & 97\,336 & 1\,168 & 865 & 136/15 & 97.14 & 0.06 & 93.56 & 96.47 & \YES & \NO & \NO &  &  & 0.53 & 4.37 & \YES\\\hline
Belarusian-HSE & 6\,541 & 229 & 291 & 46/14 & 92.39 & 0.15 & 90.28 & 95.24 & \YES & \NO & \NO &  &  & 0.00 & 3.16 & \YES\\\hline
Breton-KEB & 8\,062 & 329 & 87 & 36/13 & 93.03 & 0.15 & 91.44 & 93.95 & \YES & \NO & \NO &  &  & 0.09 & 0.89 & \YES\\\hline
Bulgarian-BTB & 124\,749 & 605 & 316 & 44/15 & 98.34 & 0.05 & 97.84 & 99.03 & \YES & \NO & \NO &  &  & 0.25 & 0.21 & \YES\\\hline
Buryat-BDT & 8\,029 & 132 & 167 & 41/12 & 90.38 & 0.23 & 87.98 & 89.94 & \YES & \NO & \NO &  &  & 0.76 & 1.35 & \NO\\\hline
Cantonese-HK & 5\,121 & 17 & 13 & 12/~1~ & 100.00 & 0.00 & 91.25 & 88.86 & \YES & \NO & \NO &  &  & 0.00 & 0.54 & \YES\\\hline
Catalan-AnCora & 427\,672 & 579 & 145 & 41/13 & 99.32 & 0.01 & 98.66 & 99.35 & \YES & \NO & \NO &  &  & 0.09 & 0.34 & \YES\\\hline
Chinese-CFL & 5\,688 & 14 & 13 & 12/~1~ & 99.76 & 0.00 & 94.21 & 93.34 & \YES & \NO & \NO & 0.00 & -0.12 & 0.00 & 3.43 & \YES\\\hline
Chinese-GSD & 98\,734 & 26 & 27 & 21/~8~ & 99.98 & 0.00 & 96.64 & 96.51 & \YES & \NO & \NO & 0.00 & 0.11 & -0.01 & 2.61 & \YES\\\hline
Coptic-Scriptorium & 17\,624 & 181 & 53 & 26/~9~ & 97.31 & 0.06 & 95.85 & 96.82 & \YES & \NO & \NO &  &  & -0.05 & -0.56 & \NO\\\hline
Croatian-SET & 157\,446 & 578 & 818 & 51/16 & 97.45 & 0.05 & 94.16 & 97.67 & \YES & \NO & \NO &  &  & 0.22 & 0.83 & \YES\\\hline
Czech-CAC & 395\,043 & 929 & 960 & 57/15 & 99.31 & 0.01 & 97.78 & 99.20 & \YES & \NO & \NO & -0.18 & -0.37 & 0.06 & 0.73 & \YES\\\hline
Czech-CLTT & 28\,649 & 229 & 350 & 48/15 & 99.22 & 0.02 & 95.42 & 98.30 & \NO & \YES & \NO & 0.11 & -0.06 & 0.37 & 1.00 & \YES\\\hline
Czech-FicTree & 133\,300 & 692 & 971 & 53/15 & 98.80 & 0.02 & 96.30 & 98.53 & \YES & \NO & \NO & 0.36 & -0.23 & 0.12 & 0.65 & \YES\\\hline
Czech-PDT & \kern-2pt1\,207\,922 & 1\,661 & 1\,123 & 57/15 & 99.37 & 0.01 & 98.02 & 99.25 & \YES & \NO & \NO & -0.05 & -0.15 & 0.04 & 0.44 & \YES\\\hline
Czech-PUD & 14\,814 & 349 & 549 & 56/15 & 98.13 & 0.03 & 94.46 & 98.14 & \NO & \YES & \NO & 2.59 & 3.94 & 0.77 & 3.84 & \YES\\\hline
Danish-DDT & 80\,964 & 426 & 128 & 41/14 & 98.30 & 0.03 & 97.76 & 98.49 & \YES & \NO & \NO &  &  & 0.44 & 0.84 & \YES\\\hline
Dutch-Alpino & 167\,187 & 631 & 43 & 33/10 & 98.45 & 0.03 & 97.59 & 98.20 & \NO & \YES & \NO & -0.03 & -0.03 & 0.12 & 0.39 & \YES\\\hline
Dutch-LassySmall & 78\,638 & 527 & 41 & 31/10 & 98.34 & 0.03 & 97.86 & 98.29 & \NO & \YES & \NO & 0.28 & 0.18 & 0.07 & 0.56 & \YES\\\hline
English-EWT & 204\,839 & 235 & 76 & 36/12 & 99.01 & 0.02 & 97.53 & 98.27 & \YES & \NO & \NO & -0.18 & -0.38 & 0.30 & 0.99 & \YES\\\hline
English-GUM & 63\,862 & 160 & 74 & 37/12 & 98.53 & 0.02 & 97.29 & 98.01 & \YES & \NO & \NO & -0.84 & -1.23 & 0.26 & 1.23 & \YES\\\hline
English-LinES & 66\,428 & 166 & 78 & 36/12 & 98.62 & 0.02 & 97.52 & 98.14 & \YES & \NO & \NO & -0.83 & -1.13 & 0.22 & 0.87 & \YES\\\hline
English-PUD & 16\,921 & 70 & 66 & 35/12 & 97.79 & 0.03 & 96.32 & 97.28 & \YES & \NO & \NO & -1.76 & -0.54 & 0.83 & 2.55 & \YES\\\hline
English-ParTUT & 39\,302 & 115 & 83 & 33/10 & 98.37 & 0.03 & 96.25 & 96.92 & \YES & \NO & \NO & -0.48 & -2.36 & 0.21 & 1.19 & \YES\\\hline
Estonian-EDT & 346\,986 & 3\,294 & 494 & 52/14 & 96.59 & 0.06 & 96.72 & 98.37 & \YES & \NO & \NO &  &  & 0.35 & 0.44 & \YES\\\hline
Faroese-OFT & 7\,994 & 297 & 234 & 36/13 & 90.30 & 0.18 & 88.28 & 94.29 & \YES & \NO & \NO &  &  & 1.64 & 2.11 & \NO\\\hline
Finnish-FTB & 127\,536 & 1\,211 & 660 & 53/12 & 96.05 & 0.08 & 96.55 & 97.98 & \YES & \NO & \NO &  &  & 0.37 & 0.44 & \YES\\\hline
Finnish-PUD & 12\,553 & 889 & 284 & 50/12 & 88.90 & 0.19 & 96.58 & 98.33 & \NO & \YES & \NO & 2.15 & 1.97 & 1.21 & 2.09 & \YES\\\hline
Finnish-TDT & 161\,582 & 2\,650 & 565 & 51/12 & 95.91 & 0.08 & 96.81 & 98.21 & \YES & \NO & \NO & -0.20 & -0.08 & 0.34 & 0.39 & \YES\\\hline
French-GSD & 320\,404 & 736 & 134 & 40/13 & 98.82 & 0.02 & 97.82 & 98.71 & \YES & \NO & \NO & 0.01 & -0.15 & 0.11 & 0.38 & \YES\\\hline
French-ParTUT & 22\,627 & 219 & 111 & 34/10 & 96.66 & 0.06 & 95.84 & 98.02 & \YES & \NO & \NO &  &  & 0.12 & 0.88 & \YES\\\hline
French-Sequoia & 56\,484 & 317 & 126 & 35/12 & 99.01 & 0.02 & 98.15 & 99.13 & \YES & \NO & \NO &  &  & 0.14 & 0.56 & \YES\\\hline
French-Spoken & 28\,182 & 208 & 13 & 12/~1~ & 98.91 & 0.02 & 98.12 & 98.15 & \YES & \NO & \NO &  &  & 0.14 & 0.25 & \YES\\\hline
Galician-CTG & 111\,034 & 160 & 14 & 13/~2~ & 98.87 & 0.02 & 98.28 & 98.12 & \YES & \NO & \NO & 0.00 & -0.04 & 0.11 & 0.29 & \YES\\\hline
Galician-TreeGal & 20\,566 & 147 & 161 & 44/13 & 98.49 & 0.03 & 95.71 & 97.63 & \YES & \NO & \NO & -0.03 & -11.16 & 0.38 & 1.27 & \YES\\\hline
German-GSD & 234\,161 & 841 & 600 & 41/12 & 97.70 & 0.05 & 89.89 & 95.64 & \YES & \NO & \NO &  &  & 0.21 & 0.79 & \YES\\\hline
Gothic-PROIEL & 44\,660 & 1\,130 & 540 & 43/13 & 95.61 & 0.09 & 90.50 & 96.39 & \NO & \NO & \YES &  &  & -0.04 & -0.13 & \NO\\\hline
Greek-GDT & 50\,567 & 1\,285 & 243 & 40/12 & 96.85 & 0.08 & 95.89 & 98.36 & \YES & \NO & \NO &  &  & 0.09 & 0.86 & \YES\\\hline
Hebrew-HTB & 129\,425 & 387 & 236 & 38/13 & 98.18 & 0.03 & 97.51 & 98.24 & \YES & \NO & \NO &  &  & 0.19 & 0.45 & \YES\\\hline
Hindi-HDTB & 281\,948 & 286 & 738 & 49/12 & 98.89 & 0.01 & 93.23 & 97.83 & \YES & \NO & \NO &  &  & 0.07 & 0.53 & \YES\\\hline
Hungarian-Szeged & 33\,463 & 329 & 427 & 59/14 & 97.45 & 0.05 & 95.22 & 98.32 & \YES & \NO & \NO &  &  & 0.25 & 2.21 & \YES\\\hline
Indonesian-GSD & 97\,213 & 65 & 129 & 27/~9~ & 99.62 & 0.01 & 92.06 & 94.75 & \YES & \NO & \NO &  &  & 0.05 & 0.67 & \YES\\\hline
Irish-IDT & 18\,996 & 476 & 163 & 37/12 & 92.06 & 0.18 & 86.41 & 91.51 & \YES & \NO & \NO &  &  & 0.00 & 0.49 & \YES\\\hline
Italian-ISDT & 239\,381 & 321 & 142 & 38/11 & 98.86 & 0.02 & 98.30 & 99.03 & \NO & \YES & \NO & 0.01 & 0.19 & 0.13 & 0.33 & \YES\\\hline
Italian-PUD & 18\,834 & 167 & 159 & 38/14 & 97.57 & 0.05 & 96.33 & 98.34 & \YES & \NO & \NO & 0.41 & -11.18 & 0.46 & 2.19 & \YES\\\hline
Italian-ParTUT & 44\,556 & 194 & 110 & 34/11 & 99.26 & 0.02 & 98.66 & 99.16 & \NO & \YES & \NO & 0.78 & 0.85 & 0.04 & 0.58 & \YES\\\hline
Italian-PoSTWITA & 99\,067 & 945 & 122 & 33/10 & 97.82 & 0.05 & 96.52 & 97.49 & \YES & \NO & \NO & -0.13 & -0.10 & 0.33 & 0.67 & \YES\\\hline
    \end{tabular}
  \end{center}
  \caption{For every corpus, its size, the number of unique lemma rules, the
  number of unique POS tags, and the number of morphological features and
  morphological categories is presented. Then the test set results of lemma
  accuracy, lemma Levenshtein, morphological accuracy and morphological F1 follow,
  using a model achieving best score on the development set. We consider the
  regular model \texttt{R}, or a model on the merged corpus \texttt{M} and
  a model without BERT embeddings \texttt{N}. Finally, we show the increase
  of the merged model to the regular model, the increase of the regular
  model to the no-BERT model, and indicate if the language is present in BERT
  training data (\texttt{BT}).}
  \label{table:results-per-corpus-I}
\end{table*}

\begin{table*}
  \begin{center}
    \small
    \tabcolsep=2pt
    \begin{tabular}{l|r|rrr||rr|rr|ccc||rr|rr|r}
      \multicolumn{1}{c|}{\multirow{2}{*}{Treebank}} &
      \multicolumn{1}{c|}{\multirow{2}{*}{Words}} &
      \multicolumn{1}{c}{\kern-2ptLemma\kern-2pt} &
      \multicolumn{1}{c}{POS} &
      \multicolumn{1}{c||}{Feats/} &
      \multicolumn{2}{c|}{Lemma} & \multicolumn{2}{c|}{Morph} &
      \multicolumn{3}{c||}{Model} &
      \multicolumn{2}{c|}{Merged $\Delta$} &
      \multicolumn{2}{c|}{BERT $\Delta$} &
      \multicolumn{1}{c}{B} \\
      &&\multicolumn{1}{c}{Rules} &
      \multicolumn{1}{c}{Tags} &
      \multicolumn{1}{c||}{/Cats} &
      \multicolumn{1}{c}{Acc} & \multicolumn{1}{c|}{Lev} & \multicolumn{1}{c}{Acc} & \multicolumn{1}{c||}{F1} &
      \multicolumn{1}{c}{R} &
      \multicolumn{1}{c}{M} &
      \multicolumn{1}{c||}{N} &
      \multicolumn{1}{c}{LAcc} &
      \multicolumn{1}{c|}{MAcc} &
      \multicolumn{1}{c}{LAcc} &
      \multicolumn{1}{c|}{MAcc} &
      \multicolumn{1}{c}{T} \\\hline\hline
Japanese-GSD & 147\,897 & 104 & 13 & 12/~1~ & 99.65 & 0.00 & 98.14 & 97.91 & \NO & \YES & \NO & 0.03 & -0.02 & -0.01 & 0.23 & \YES\\\hline
Japanese-Modern & 11\,556 & 44 & 14 & 12/~2~ & 98.67 & 0.01 & 96.80 & 96.87 & \YES & \NO & \NO & -0.14 & -0.20 & -0.20 & 0.33 & \YES\\\hline
Japanese-PUD & 21\,650 & 51 & 12 & 11/~1~ & 99.77 & 0.00 & 99.32 & 99.25 & \NO & \YES & \NO & 0.64 & 0.99 & 0.19 & 0.34 & \YES\\\hline
Komi Zyrian-IKDP & 847 & 85 & 114 & 40/11 & 85.94 & 0.25 & 76.56 & 86.19 & \NO & \YES & \NO & 2.35 & 0.78 & 1.56 & 5.47 & \NO\\\hline
Komi Zyrian-Lattice & 1\,653 & 58 & 184 & 46/13 & 87.36 & 0.28 & 73.63 & 85.36 & \NO & \YES & \NO & 1.10 & 3.85 & 0.55 & 1.10 & \NO\\\hline
Korean-GSD & 64\,311 & 1\,470 & 13 & 12/~1~ & 93.77 & 0.12 & 96.47 & 95.92 & \YES & \NO & \NO & -0.63 & -5.78 & 0.59 & 1.08 & \YES\\\hline
Korean-Kaist & 280\,494 & 3\,137 & 13 & 12/~1~ & 95.65 & 0.07 & 97.31 & 96.98 & \YES & \NO & \NO & -0.03 & -0.23 & 0.17 & 0.31 & \YES\\\hline
Korean-PUD & 13\,306 & 9 & 109 & 29/11 & 99.07 & 0.01 & 94.06 & 96.23 & \YES & \NO & \NO & -10.90 & -16.72 & 0.06 & 2.67 & \YES\\\hline
Kurmanji-MG & 8\,077 & 275 & 148 & 41/14 & 94.71 & 0.10 & 85.57 & 91.52 & \YES & \NO & \NO &  &  & 0.09 & 0.80 & \NO\\\hline
Latin-ITTB & 281\,652 & 726 & 539 & 46/12 & 98.99 & 0.02 & 96.85 & 98.49 & \YES & \NO & \NO & 0.10 & 0.02 & 0.04 & 0.13 & \YES\\\hline
Latin-PROIEL & 160\,257 & 1\,555 & 872 & 48/13 & 97.28 & 0.06 & 92.40 & 97.23 & \NO & \YES & \NO & 0.07 & 0.12 & 0.02 & 0.06 & \YES\\\hline
Latin-Perseus & 23\,339 & 879 & 427 & 43/11 & 93.32 & 0.14 & 86.97 & 94.28 & \NO & \YES & \NO & 2.99 & 2.60 & 0.17 & 0.48 & \YES\\\hline
Latvian-LVTB & 121\,760 & 677 & 644 & 49/15 & 97.22 & 0.05 & 95.48 & 97.74 & \YES & \NO & \NO &  &  & 0.04 & 0.22 & \YES\\\hline
Lithuanian-HSE & 4\,301 & 209 & 337 & 45/13 & 87.27 & 0.26 & 82.34 & 89.59 & \YES & \NO & \NO &  &  & 0.68 & 2.88 & \YES\\\hline
Marathi-UFAL & 3\,055 & 236 & 222 & 45/11 & 76.42 & 0.66 & 67.21 & 79.00 & \YES & \NO & \NO &  &  & -0.54 & 0.54 & \YES\\\hline
Naija-NSC & 10\,280 & 7 & 13 & 12/~1~ & 99.93 & 0.00 & 96.28 & 95.06 & \YES & \NO & \NO &  &  & 0.00 & 0.45 & \NO\\\hline
North Sami-Giella & 21\,380 & 1\,019 & 314 & 51/13 & 92.18 & 0.16 & 91.78 & 94.96 & \YES & \NO & \NO &  &  & -0.25 & 0.04 & \NO\\\hline
Norwegian-Bokmaal & 248\,922 & 445 & 142 & 42/14 & 99.14 & 0.01 & 97.88 & 98.77 & \YES & \NO & \NO &  &  & 0.03 & 0.31 & \YES\\\hline
Norwegian-Nynorsk & 241\,028 & 478 & 138 & 40/12 & 98.96 & 0.02 & 97.48 & 98.49 & \NO & \YES & \NO & 0.05 & -0.01 & 0.11 & 0.43 & \YES\\\hline
Norwegian-NynorskLIA & 10\,843 & 111 & 100 & 37/14 & 98.15 & 0.03 & 96.30 & 97.25 & \NO & \YES & \NO & 0.35 & 0.35 & -0.07 & 0.43 & \YES\\\hline
\vtop to 15pt{\vbox{\hbox{Old Church Slavonic}\kern4pt}\kern-2pt\hbox{~~~~-PROIEL}} & 45\,894 & 1\,796 & 726 & 48/13 & 94.71 & 0.11 & 92.92 & 97.06 & \YES & \NO & \NO &  &  & -0.07 & 0.13 & \NO\\\hline
Persian-Seraji & 122\,574 & 772 & 104 & 31/10 & 96.86 & 0.16 & 98.30 & 98.67 & \YES & \NO & \NO &  &  & 0.27 & 0.60 & \YES\\\hline
Polish-LFG & 104\,730 & 819 & 609 & 50/14 & 97.79 & 0.04 & 96.42 & 98.55 & \YES & \NO & \NO & -0.07 & -0.89 & 0.18 & 0.72 & \YES\\\hline
Polish-SZ & 66\,430 & 695 & 717 & 51/15 & 97.45 & 0.04 & 94.61 & 97.89 & \YES & \NO & \NO &  &  & 0.34 & 1.94 & \YES\\\hline
Portuguese-Bosque & 180\,773 & 402 & 247 & 43/12 & 98.70 & 0.02 & 96.09 & 98.18 & \YES & \NO & \NO & -4.84 & -5.27 & 0.19 & 0.76 & \YES\\\hline
Portuguese-GSD & 255\,690 & 175 & 19 & 17/~5~ & 99.07 & 0.05 & 98.88 & 98.96 & \YES & \NO & \NO & -2.49 & -0.62 & 0.19 & 0.49 & \YES\\\hline
Romanian-Nonstandard & 156\,320 & 2\,094 & 288 & 45/14 & 96.78 & 0.06 & 94.62 & 97.27 & \NO & \YES & \NO & 0.02 & -0.02 & 0.23 & 0.38 & \YES\\\hline
Romanian-RRT & 174\,747 & 678 & 254 & 47/14 & 98.50 & 0.03 & 97.97 & 98.68 & \YES & \NO & \NO & -0.03 & -0.05 & 0.15 & 0.30 & \YES\\\hline
Russian-GSD & 79\,989 & 553 & 668 & 47/14 & 97.93 & 0.04 & 94.38 & 97.64 & \YES & \NO & \NO &  &  & 0.89 & 4.05 & \YES\\\hline
Russian-PUD & 15\,433 & 309 & 525 & 46/15 & 94.69 & 0.09 & 90.24 & 96.45 & \YES & \NO & \NO & 0.96 & -7.25 & 1.88 & 7.14 & \YES\\\hline
Russian-SynTagRus & 886\,711 & 1\,744 & 678 & 48/13 & 98.92 & 0.02 & 98.05 & 99.05 & \YES & \NO & \NO & -0.04 & -0.08 & 0.24 & 1.02 & \YES\\\hline
Russian-Taiga & 16\,762 & 434 & 383 & 47/13 & 95.33 & 0.10 & 89.36 & 94.74 & \NO & \YES & \NO & 2.64 & -0.40 & 2.14 & 7.02 & \YES\\\hline
Sanskrit-UFAL & 1\,450 & 244 & 232 & 54/14 & 64.82 & 0.89 & 50.25 & 68.99 & \NO & \NO & \YES &  &  & 0.00 & 0.00 & \NO\\\hline
Serbian-SET & 68\,933 & 359 & 421 & 39/12 & 98.27 & 0.03 & 96.68 & 98.40 & \YES & \NO & \NO &  &  & 0.50 & 1.04 & \YES\\\hline
Slovak-SNK & 85\,257 & 598 & 830 & 52/15 & 97.49 & 0.04 & 94.96 & 97.96 & \YES & \NO & \NO &  &  & 0.28 & 1.30 & \YES\\\hline
Slovenian-SSJ & 112\,136 & 369 & 744 & 52/15 & 98.84 & 0.02 & 96.99 & 98.59 & \NO & \YES & \NO & 0.06 & 0.14 & 0.28 & 1.03 & \YES\\\hline
Slovenian-SST & 23\,759 & 214 & 473 & 49/14 & 97.74 & 0.05 & 93.52 & 95.96 & \NO & \YES & \NO & 1.53 & 1.68 & 0.35 & 0.66 & \YES\\\hline
Spanish-AnCora & 439\,925 & 594 & 173 & 42/13 & 99.48 & 0.01 & 98.63 & 99.28 & \YES & \NO & \NO & -0.24 & -0.19 & 0.17 & 0.35 & \YES\\\hline
Spanish-GSD & 345\,545 & 310 & 239 & 52/14 & 99.31 & 0.01 & 95.67 & 97.97 & \YES & \NO & \NO & -0.27 & -0.50 & 0.05 & 0.26 & \YES\\\hline
Swedish-LinES & 63\,365 & 332 & 135 & 38/11 & 98.05 & 0.04 & 94.66 & 97.47 & \YES & \NO & \NO & -0.19 & -0.56 & -0.02 & 0.57 & \YES\\\hline
Swedish-PUD & 14\,952 & 171 & 94 & 36/11 & 95.85 & 0.07 & 95.39 & 97.25 & \NO & \YES & \NO & 0.00 & -0.04 & 0.74 & 2.21 & \YES\\\hline
Swedish-Talbanken & 77\,238 & 291 & 119 & 38/11 & 98.60 & 0.02 & 97.84 & 98.87 & \YES & \NO & \NO & -0.16 & -0.21 & 0.08 & 0.69 & \YES\\\hline
Tagalog-TRG & 230 & 19 & 33 & 31/11 & 91.89 & 0.30 & 91.89 & 95.04 & \YES & \NO & \NO &  &  & -5.41 & 0.00 & \YES\\\hline
Tamil-TTB & 7\,634 & 99 & 172 & 47/13 & 96.65 & 0.07 & 91.85 & 96.11 & \YES & \NO & \NO &  &  & 1.45 & 2.01 & \YES\\\hline
Turkish-IMST & 46\,417 & 211 & 985 & 56/13 & 96.84 & 0.06 & 92.83 & 96.60 & \YES & \NO & \NO & -0.31 & -0.83 & 0.15 & 0.82 & \YES\\\hline
Turkish-PUD & 13\,380 & 103 & 503 & 62/13 & 88.02 & 0.30 & 88.69 & 95.07 & \YES & \NO & \NO & 1.23 & -2.39 & 0.33 & 2.67 & \YES\\\hline
Ukrainian-IU & 93\,264 & 629 & 597 & 49/14 & 97.84 & 0.04 & 95.40 & 97.93 & \YES & \NO & \NO &  &  & 0.25 & 1.51 & \YES\\\hline
Upper Sorbian-UFAL & 8\,959 & 222 & 417 & 49/14 & 93.46 & 0.11 & 87.11 & 93.71 & \YES & \NO & \NO &  &  & 0.56 & 2.89 & \NO\\\hline
Urdu-UDTB & 110\,682 & 448 & 740 & 49/12 & 97.10 & 0.05 & 80.95 & 93.44 & \YES & \NO & \NO &  &  & 0.34 & 0.82 & \YES\\\hline
Vietnamese-VTB & 35\,237 & 51 & 13 & 11/~2~ & 99.91 & 0.00 & 93.49 & 92.71 & \YES & \NO & \NO &  &  & 0.05 & 1.06 & \YES\\\hline
Yoruba-YTB & 2\,158 & 3 & 29 & 19/~4~ & 97.67 & 0.02 & 91.86 & 92.66 & \YES & \NO & \NO &  &  & 0.00 & 0.00 & \YES\\\hline
    \end{tabular}
  \end{center}
  \caption{For every corpus, its size, the number of unique lemma rules, the
  number of unique POS tags, and the number of morphological features and
  morphological categories is presented. Then the test set results of lemma
  accuracy, lemma Levenshtein, morphological accuracy and morphological F1 follow,
  using a model achieving best score on the development set. We consider the
  regular model \texttt{R}, or a model on the merged corpus \texttt{M} and
  a model without BERT embeddings \texttt{N}. Finally, we show the increase
  of the merged model to the regular model, the increase of the regular
  model to the no-BERT model, and indicate if the language is present in BERT
  training data (\texttt{BT}).}
  \label{table:results-per-corpus-II}
\end{table*}

\subsection{Ablation Experiments}

The effect of pretrained word embeddings, BERT contextualized embeddings and
regularization with morphological features is evaluated in
Table~\ref{table:ablations}. Even the baseline model without any of the mentioned enhancements achieves
relatively high performance and would place third in both lemmatization and
tagging accuracy (when not considering our competition entry).

Pretrained word
embeddings improve the performance of both the lemmatizer and the tagger by
a substantial margin. For comparison with the embeddings we trained on CoNLL
2017 UD Shared Task plain texts, we also evaluate the embeddings provided
by \citet{crawlvectors}, which achieve only slightly lower performance than our
embeddings -- we presume the difference is caused mostly by different
tokenization, given that the training data comes from Wikipedia
and CommonCrawl in both cases.

BERT contextualized embeddings further considerably improve POS tagging performance,
and have minor impact on lemmatization improvement.

When used in isolation, the regularization with morphological categories
provides quite considerable gain for both lemmatization and tagging, nearly comparable
to the effect of adding precomputed word embeddings. Combining all the
enhancements together then produces a model with the highest performance.

\subsection{Model Combinations}

For every corpus, we consider three model configurations -- a regular model,
then a model trained on the merged corpora of a corresponding language, and
a model without BERT embeddings (which we consider since even if BERT embeddings can
be computed for any language, the results might be misleading if the language
was not present in the BERT training data). For every model configuration, we
train three models using different random initialization.

The test set results of choosing the best model configuration on a development
set are provided in Table~\ref{table:combination}. Employing the merged
model in addition to the regular model increases the performance slightly, and
the introduction of no-BERT model results in minimal gains. Finally, ensembling the
models of a same configuration provides the highest performance.

As discussed in Section~\ref{section:model_ensembling}, our competition entry
selected the ensemble using arbitrary subset of all the nine models which achieved
best performance on the development data. This choice resulted in overfitting
on POS tag prediction, with results worse than no ensembling.

\subsection{Detailed Results}

Tables~\ref{table:results-per-corpus-I}~and~\ref{table:results-per-corpus-II}
present detailed results of our best system from Table~\ref{table:combination}.
Note that while this system is not our competition entry, it utilizes the
same models as the competition entry, only combined in a different way.
Furthermore, because one model configuration was chosen for every language,
we can examine which configuration performed best, and quantify what the exact
effect of corpora merging and BERT embeddings are.

\section{Conclusions}
\label{section:conclusions}

We described our system which participated in the \sigmorphon, Task 2:
contextual morphological analysis and lemmatization, which placed first in
lemmatization and closely second in morphological analysis. The contributed
architecture is a modified \udpipe with three improvements: addition of
pretrained contextualized BERT embeddings, regularization with morphological
categories and corpora merging in some languages. We described these improvements and published
the related ablation experiment results.

\section*{Acknowledgements}

The work described herein has been supported by OP VVV VI LINDAT/CLARIN project
of the Ministry of Education, Youth and Sports of the Czech Republic (project
CZ.02.1.01/0.0/0.0/16\_013/0001781) and it has been supported and has been
using language resources developed by the LINDAT/CLARIN project of the Ministry
of Education, Youth and Sports of the Czech Republic (project LM2015071).

\bibliographystyle{acl_natbib}
\bibliography{sigmorphon2019}

\begin{thebibliography}{18}
\expandafter\ifx\csname natexlab\endcsname\relax\def\natexlab#1{#1}\fi

\bibitem[{Bojanowski et~al.(2017)Bojanowski, Grave, Joulin, and
  Mikolov}]{FastText}
Piotr Bojanowski, Edouard Grave, Armand Joulin, and Tomas Mikolov. 2017.
\newblock \href {http://aclweb.org/anthology/Q17-1010} {{Enriching Word Vectors
  with Subword Information}}.
\newblock \emph{Transactions of the Association for Computational Linguistics},
  5:135--146.

\bibitem[{Che et~al.(2018)Che, Liu, Wang, Zheng, and Liu}]{udst18:harbin}
Wanxiang Che, Yijia Liu, Yuxuan Wang, Bo~Zheng, and Ting Liu. 2018.
\newblock \href {http://www.aclweb.org/anthology/K18-2005} {Towards better {UD}
  parsing: Deep contextualized word embeddings, ensemble, and treebank
  concatenation}.
\newblock In \emph{Proceedings of the {CoNLL} 2018 Shared Task: Multilingual
  Parsing from Raw Text to Universal Dependencies}, pages 55--64, Brussels,
  Belgium. Association for Computational Linguistics.

\bibitem[{Cho et~al.(2014)Cho, van Merrienboer, Bahdanau, and Bengio}]{Cho2014}
KyungHyun Cho, Bart van Merrienboer, Dzmitry Bahdanau, and Yoshua Bengio. 2014.
\newblock On the {P}roperties of {N}eural {M}achine {T}ranslation:
  {E}ncoder-{D}ecoder {A}pproaches.
\newblock \emph{CoRR}.

\bibitem[{Devlin et~al.(2018)Devlin, Chang, Lee, and Toutanova}]{BERT}
Jacob Devlin, Ming-Wei Chang, Kenton Lee, and Kristina Toutanova. 2018.
\newblock \href {https://arxiv.org/abs/1810.04805} {{BERT: Pre-training of Deep
  Bidirectional Transformers for Language Understanding}}.

\bibitem[{Fares et~al.(2018)Fares, Oepen, {\O}vrelid, Bj\"orne, and
  Johansson}]{EPE2018}
Murhaf Fares, Stephan Oepen, Lilja {\O}vrelid, Jari Bj\"orne, and Richard
  Johansson. 2018.
\newblock \href {http://aclweb.org/anthology/K18-2002} {{The 2018 Shared Task
  on Extrinsic Parser Evaluation: On the Downstream Utility of English
  Universal Dependency Parsers}}.
\newblock In \emph{Proceedings of the CoNLL 2018 Shared Task: Multilingual
  Parsing from Raw Text to Universal Dependencies}, pages 22--33. Association
  for Computational Linguistics.

\bibitem[{Grave et~al.(2018)Grave, Bojanowski, Gupta, Joulin, and
  Mikolov}]{crawlvectors}
Edouard Grave, Piotr Bojanowski, Prakhar Gupta, Armand Joulin, and Tomas
  Mikolov. 2018.
\newblock Learning word vectors for 157 languages.
\newblock In \emph{Proceedings of the International Conference on Language
  Resources and Evaluation (LREC 2018)}.

\bibitem[{Graves and Schmidhuber(2005)}]{Graves2005}
Alex Graves and J{\"{u}}rgen Schmidhuber. 2005.
\newblock Framewise phoneme classification with bidirectional lstm and other
  neural network architectures.
\newblock \emph{Neural Networks}, pages 5--6.

\bibitem[{Hochreiter and Schmidhuber(1997)}]{Hochreiter:1997:LSTM}
Sepp Hochreiter and J{\"{u}}rgen Schmidhuber. 1997.
\newblock Long {S}hort-{T}erm {M}emory.
\newblock \emph{Neural Comput.}, 9(8):1735--1780.

\bibitem[{Kondratyuk et~al.(2018)Kondratyuk, Gavenciak, Straka, and
  Hajic}]{LemmaTag}
Daniel Kondratyuk, Tomas Gavenciak, Milan Straka, and Jan Hajic. 2018.
\newblock \href {https://aclanthology.info/papers/D18-1532/d18-1532} {Lemmatag:
  Jointly tagging and lemmatizing for morphologically rich languages with
  brnns}.
\newblock In \emph{Proceedings of the 2018 Conference on Empirical Methods in
  Natural Language Processing, Brussels, Belgium, October 31 - November 4,
  2018}, pages 4921--4928. Association for Computational Linguistics.

\bibitem[{Ling et~al.(2015)Ling, Lu{\'{i}}s, Marujo, Astudillo, Amir, Dyer,
  Black, and Trancoso}]{Ling2015}
Wang Ling, Tiago Lu{\'{i}}s, Lu{\'{i}}s Marujo, Ram{\'{o}}n~Fernandez
  Astudillo, Silvio Amir, Chris Dyer, Alan~W. Black, and Isabel Trancoso. 2015.
\newblock Finding {F}unction in {F}orm: {C}ompositional {C}haracter {M}odels
  for {O}pen {V}ocabulary {W}ord {R}epresentation.
\newblock \emph{CoRR}.

\bibitem[{McCarthy et~al.(2018)McCarthy, Silfverberg, Cotterell, Hulden, and
  Yarowsky}]{mccarthy-etal-2018-marrying}
Arya~D. McCarthy, Miikka Silfverberg, Ryan Cotterell, Mans Hulden, and David
  Yarowsky. 2018.
\newblock \href {https://www.aclweb.org/anthology/W18-6011} {Marrying
  {U}niversal {D}ependencies and {U}niversal {M}orphology}.
\newblock In \emph{Proceedings of the Second Workshop on Universal Dependencies
  ({UDW} 2018)}, pages 91--101, Brussels, Belgium. Association for
  Computational Linguistics.

\bibitem[{Nivre et~al.(2016)Nivre, de~Marneffe, Ginter, Goldberg, Haji{\v{c}},
  Manning, McDonald, Petrov, Pyysalo, Silveira, Tsarfaty, and Zeman}]{ud}
Joakim Nivre, Marie-Catherine de~Marneffe, Filip Ginter, Yoav Goldberg, Jan
  Haji{\v{c}}, Christopher Manning, Ryan McDonald, Slav Petrov, Sampo Pyysalo,
  Natalia Silveira, Reut Tsarfaty, and Daniel Zeman. 2016.
\newblock {Universal Dependencies} v1: A multilingual treebank collection.
\newblock In \emph{Proceedings of the 10th International Conference on Language
  Resources and Evaluation ({LREC} 2016)}, pages 1659--1666, Portorož,
  Slovenia. European Language Resources Association.

\bibitem[{Peters et~al.(2018)Peters, Neumann, Iyyer, Gardner, Clark, Lee, and
  Zettlemoyer}]{Peters2018}
Matthew Peters, Mark Neumann, Mohit Iyyer, Matt Gardner, Christopher Clark,
  Kenton Lee, and Luke Zettlemoyer. 2018.
\newblock {Deep Contextualized Word Representations}.
\newblock In \emph{Proceedings of the 2018 Conference of the North American
  Chapter of the Association for Computational Linguistics: Human Language
  Technologies, Volume 1 (Long Papers)}, pages 2227--2237. Association for
  Computational Linguistics.

\bibitem[{Straka(2018)}]{UDPipe2.0}
Milan Straka. 2018.
\newblock {UDPipe 2.0 Prototype at CoNLL 2018 UD Shared Task}.
\newblock In \emph{Proceedings of CoNLL 2018: The SIGNLL Conference on
  Computational Natural Language Learning}, pages 197--207, Stroudsburg, PA,
  USA. Association for Computational Linguistics.

\bibitem[{Sylak-Glassman(2016)}]{UniMorph}
John Sylak-Glassman. 2016.
\newblock The composition and use of the universal morphological feature schema
  (unimorph schema).

\bibitem[{Vaswani et~al.(2017)Vaswani, Shazeer, Parmar, Uszkoreit, Jones,
  Gomez, Kaiser, and Polosukhin}]{vaswani:2017}
Ashish Vaswani, Noam Shazeer, Niki Parmar, Jakob Uszkoreit, Llion Jones,
  Aidan~N. Gomez, Lukasz Kaiser, and Illia Polosukhin. 2017.
\newblock \href {http://arxiv.org/abs/1706.03762} {Attention is all you need}.
\newblock \emph{CoRR}, abs/1706.03762.

\bibitem[{Zeman et~al.(2018)Zeman, Ginter, Haji{\v{c}}, Nivre, Popel, and
  Straka}]{CoNLL2018}
Daniel Zeman, Filip Ginter, Jan Haji{\v{c}}, Joakim Nivre, Martin Popel, and
  Milan Straka. 2018.
\newblock {CoNLL 2018 Shared Task: Multilingual Parsing from Raw Text to
  Universal Dependencies}.
\newblock In \emph{{Proceedings of the CoNLL 2018 Shared Task: Multilingual
  Parsing from Raw Text to Universal Dependencies}}, Brussels, Belgium.
  Association for Computational Linguistics.

\bibitem[{Zeman et~al.(2017)Zeman, Popel, Straka, Haji{\v{c}}, Nivre, Ginter
  et~al.}]{CoNLL2017}
Daniel Zeman, Martin Popel, Milan Straka, Jan Haji{\v{c}}, Joakim Nivre, Filip
  Ginter, et~al. 2017.
\newblock {CoNLL 2017 Shared Task: Multilingual Parsing from Raw Text to
  Universal Dependencies}.
\newblock In \emph{{Proceedings of the CoNLL 2017 Shared Task: Multilingual
  Parsing from Raw Text to Universal Dependencies}}, pages 1--19, Vancouver,
  Canada. Association for Computational Linguistics.

\end{thebibliography}

\end{document}